\documentclass[]{ceurart}

\usepackage{listings}
\usepackage[T1]{fontenc}
\usepackage{graphicx}
\usepackage{amsmath}
\usepackage{multirow}
\usepackage{bm}
\usepackage{color}
\usepackage{makecell}

\usepackage{amssymb,amsbsy}
\usepackage{algpseudocode}
\usepackage{algorithm}

\usepackage{pifont}
\newcommand{\cmark}{\ding{51}} % checkmark

\begin{document}

%% Rights management information.
%% CC-BY is default license.
\copyrightyear{2025}
\copyrightclause{Copyright for this paper by its authors.
  Use permitted under Creative Commons License Attribution 4.0
  International (CC BY 4.0).}

%% This command is for the conference information
\conference{EvalLAC'25: 2nd Workshop on Automatic Evaluation of Learning and Assessment Content, July 26, 2025, Palermo, Italy}

\title{Leveraging AI Graders for Missing Score Imputation to Achieve Accurate Ability Estimation in Constructed-Response Tests}

\author[1]{Masaki Uto}[%
orcid=0000-0002-9330-5158,
email=uto@ai.lab.uec.ac.jp,
url=https://sites.google.com/site/utomasakieng/
]
\cormark[1]
\address[1]{The University of Electro-Communications,
  1-5-1 Chofugaoka, Chofu, Tokyo, Japan}

\author[1]{Yuma Ito}

\cortext[1]{Corresponding author.}

\begin{abstract}
Evaluating the abilities of learners is a fundamental objective in the field of education. In particular, there is an increasing need to assess higher-order abilities such as expressive skills and logical thinking. Constructed-response tests such as short-answer and essay-based questions have become widely used as a method to meet this demand. Although these tests are effective, they require substantial manual grading, making them both labor-intensive and costly. Item response theory (IRT) provides a promising solution by enabling the estimation of ability from incomplete score data, where human raters grade only a subset of answers provided by learners across multiple test items. However, the accuracy of ability estimation declines as the proportion of missing scores increases. Although data augmentation techniques for imputing missing scores have been explored in order to address this limitation, they often struggle with inaccuracy for sparse or heterogeneous data. To overcome these challenges, this study proposes a novel method for imputing missing scores by leveraging automated scoring technologies for accurate IRT-based ability estimation. The proposed method achieves high accuracy in ability estimation while markedly reducing manual grading workload.
\end{abstract}


\maketitle

\section{Introduction}\label{sec:intro}

Evaluating the abilities of learners is a critical component of various educational assessments, including entrance and qualification exams, as well as in-class assessments. Ability estimation is also essential for educational applications such as personalized learning support systems, including intelligent tutoring, and adaptive learning platforms because they generally require ability estimation to provide optimal recommendations for learning strategies, content, and other interventions tailored to the ability of each learner~\cite{Hajjioui2024,Khine2024,Shen2024,Tomikawa2024}.

Objective tests, typically consisting of multiple-choice questions, have been widely adopted as a popular approach for ability estimation in educational settings owing to their scalability and ease of implementation. However, modern education increasingly emphasizes the importance of 21st-century skills such as expressive abilities and critical thinking~\cite{Schendel2017,John2015,Lydia2014,Mislevy2018,Murtonen2019}. To effectively assess such abilities, constructed-response tests, including short-answer and essay-type questions, have gained increasing attention. These tests, however, necessitate substantial 
manual grading, which makes them both labor-intensive and costly~\cite{Amorim2018AutomatedES,ijcai2019-AesSurvey,Leckie2011DRF,UtoOkanoIEEE}.

Item response theory (IRT)~\cite{lord1980}, a statistical method well-established in the fields of educational and psychological measurement, offers a promising solution for estimating ability using incomplete score data, where human raters grade only a subset of learner answers across multiple test items, as exemplified in Table~\ref{table:ex_design}. IRT typically estimates learner ability by maximizing the likelihood of observed scores based on IRT models, which define the probability of score observations as a function of learner ability and item characteristic parameters. This allows IRT to be easily applied to incomplete score data by calculating the likelihood while excluding missing scores~\cite{zero,HIRT2016v2,PIRT2010}. This feature is particularly advantageous for achieving ability estimation while reducing manual grading workload. However, the accuracy of ability estimation decreases as the proportion of missing scores increases.

\begin{table}[t]
\centering
\caption{Examples of incomplete data, where cells marked with a \cmark indicate scored answers and blank cells indicate missing scores.}\label{table:ex_design}
\begin{minipage}{0.48\linewidth}
\tabcolsep = 2.5pt
\centering
\begin{tabular}{cccccccccc}\hline
&\multicolumn{8}{c}{Learner} \\ 
\cline{2-10}
& 1 & 2 & 3 & 4  & 5  & 6  & 7 & 8 & $\cdots$ \\
\hline
Test item~1 & \cmark & & \cmark & \cmark & & \cmark & \cmark & & $\cdots$\\
Test item~2 & \cmark & \cmark & & \cmark & \cmark & & \cmark & \cmark & $\cdots$\\
Test item~3 & & \cmark & \cmark & & \cmark & \cmark & & \cmark & $\cdots$\\
\hline
\end{tabular}
\end{minipage}%
\hfill
\begin{minipage}{0.48\linewidth}
\tabcolsep = 2.5pt
\centering
\begin{tabular}{cccccccccc}\hline
&\multicolumn{8}{c}{Learner} \\ 
\cline{2-10}
& 1 & 2 & 3 & 4  & 5  & 6  & 7 & 8 & $\cdots$ \\
\hline
Test item~1 & \cmark & & \cmark & \cmark & & & \cmark & & $\cdots$\\
Test item~2 & \cmark & \cmark & & & \cmark & & \cmark & \cmark & $\cdots$\\
Test item~3 & & \cmark & \cmark & & & \cmark & & \cmark & $\cdots$\\
\hline
\end{tabular}
\end{minipage}
\end{table}

A possible strategy for addressing this problem is data augmentation through imputation of missing scores~\cite{Barrabes2024,Zhao2023}. Simple methods such as mean or mode imputation~\cite{Little2019} are widely recognized but fail to effectively capture the underlying complex patterns within data. More advanced approaches such as statistical model-based and machine learning-based methods~\cite{Bertsimas2018,Bras2007,Husson2019,Troyanskaya2001,Wang2022,Zhang2008} aim to model such underlying patterns to predict missing scores for more accurate imputation. However, when the assumed model does not fit the target data or when the missing rate is very high, the resulting imputations often lack accuracy. Moreover, conventional imputation methods generally rely on the assumption that all data, including both observed and missing data, follow a single underlying pattern modeled as a specific data-generation process, while real-world constructed-response tests frequently violate this assumption. For instance, proficient learners could achieve lower scores due to inattention, while less proficient learners could obtain higher scores due to compatibility between the learner and the test item or other chance factors. These issues suggest that conventional imputation methods may lack the robustness for sparsity and heterogeneous data, making them unsuitable for achieving accurate ability estimation while substantially reducing manual scoring workload.

To overcome these limitations, this study proposes a novel method for imputing missing scores by leveraging automated scoring technologies~\cite{ijcai2019-AesSurvey,Chaudhari2024,Kortemeyer2024,Li2024,Misgna2025,Uto2021bhmkAES} for accurate IRT-based ability estimation in constructed-response tests. Specifically, the approach begins by developing neural automated scoring models trained on a subset of manually scored responses for each test item, or by employing zero-shot scoring models using large language models (LLMs). These models are then used to predict missing scores, generating a complete dataset. The augmented dataset is subsequently used to estimate learner ability by using IRT models. The proposed method offers several key advantages:
\begin{enumerate}
    \item More robust imputation is achieved, even for heterogeneous data, by using learner answer text directly to predict missing scores without the need to model underlying patterns of score data.
    \item Recent scoring models based on pre-trained neural models are expected to enable accurate imputations from a relatively small subset of score data, as demonstrated in recent automated scoring studies~\cite{Kortemeyer2024,do2023,Jiang2024,Ridley2021AAAI,Shibata2024}. This facilitates accurate ability estimation while markedly reducing reliance on human grading.
\end{enumerate}

Through empirical evaluation using real-world datasets, this study demonstrates that the proposed method achieves markedly higher accuracy in ability estimation than do conventional approaches, even with high missing ratios. While the proposed method is based on a relatively simple idea, its effectiveness in greatly improving the accuracy of ability estimation directly contributes to enhancing various educational applications, as outlined above.

\section{Task Settings and Objective}

The objective of this study is to estimate learner ability based on a collection of scores assigned to their answers for multiple constructed-response test items. This collection of scores is defined as follows:
\begin{equation}
\mathbf{U}  = \{ u_{ij} \in {\cal K} \cup \{-1\} \mid j \in {\cal J}, i \in {\cal I} \},
\end{equation}
where $u_{ij}$ indicates the score assigned to the answer of learner $j \in {\cal J}$ for item $i \in {\cal I}$, and ${\cal I}$ and ${\cal J}$ represent the sets of items and learners, respectively. Furthermore, ${\cal K} = \{1, 2, \ldots, K\}$ represents the set of score categories, where $K$ indicates the number of categories, and $u_{ij} = -1$ indicates missing data.

A common approach to estimating learner ability is to calculate the average or total score for each learner. However, such simple methods are not suitable for datasets with missing data, such as those exemplified in Table~\ref{table:ex_design}. This limitation arises because item characteristics such as difficulty and discrimination vary among items, causing average or total scores to depend heavily on which items each learner is graded on~\cite{Uto2018Heliyon,Uto2020bhmk}. This property is not suitable for our objective, which is to accurately estimate learner ability from score data with substantial missing values to reduce assessment workload. To address this limitation, we utilize IRT, a robust framework for estimating learner ability from incomplete data.

\section{Item Response Theory}\label{sec:irt}

IRT~\cite{lord1980} is a test theory based on statistical models that has been widely employed for ability estimation and item analysis in various educational tests. IRT estimates learner ability by considering the characteristics of test items, such as item difficulty and discrimination. This is done using probabilistic models known as IRT models, which define the probability of score observations as a function of learner ability and item characteristics. 

Among the various IRT models, the generalized partial credit model (GPCM)~\cite{MurakiGPCM1997} is a representative model particularly suited for Likert-scale polytomous score data, as assumed in this study. The GPCM defines the probability that learner $j$ receives score $k$ for constructed-response test item $i$ as 
\begin{equation}
P(u_{ij} = k)= \frac{\exp \sum_{m=1}^{k}\left[\alpha_i (\theta_j-\beta_{i}-d_{im}) \right]}{\sum_{l=1}^{K} \exp \sum_{m=1}^{l}\left[\alpha_i (\theta_j-\beta_{i}-d_{im}) \right]}, \label{eq:gpcm}
\end{equation}
where 
$\theta_j$ represents the latent ability of learner $j$, 
$\alpha_i$ is the discrimination parameter for item $i$, 
$\beta_{i}$ is the difficulty parameter for item $i$, 
and $d_{im}$ is the step difficulty parameter representing the difficulty of transitioning between scores $m-1$ and $m$ for the item. 
For model identification, $d_{i1}=0$ and $\sum_{m=2}^{K} d_{im} = 0$ are assumed.

The parameters of learner ability and item characteristics are estimated from a score collection $\mathbf{U}$, typically by maximizing the following log-likelihood function:
\begin{equation}
\log \mathcal{L}  = \sum_{i\in {\cal I}} \sum_{j\in {\cal J}} z_{ij} \log P(u_{ij}),
\end{equation}
where $z_{ij}$ is a dummy variable that equals $0$ if $u_{ij} = -1$ and $1$ otherwise. As evident from this equation, IRT can estimate parameters, including ability, from datasets with missing scores by calculating the likelihood while excluding missing scores~\cite{zero,HIRT2016v2,PIRT2010}. Furthermore, IRT generally provides more accurate ability estimates compared with methods based on simple averages or total scores because it accounts for the characteristics of test items during the estimation process~\cite{Uto2018Heliyon,Uto2020bhmk}. However, even with the IRT approach, the accuracy of ability estimation diminishes as the proportion of missing scores increases. A common strategy to address this limitation is the application of data augmentation techniques to impute missing scores.

\section{Data Augmentation}\label{sec:imputation}

There are various methods for imputation-based data augmentation. Simple approaches include mean or mode imputation, in which missing scores are replaced with averages or the most frequent scores~\cite{Little2019}. Although computationally efficient, these methods often produce biased estimates because they fail to effectively capture the underlying patterns within the data.

More advanced approaches such as statistical model-based and machine learning-based methods aim to predict missing scores by modeling such underlying patterns for more accurate imputation~\cite{Bertsimas2018,Bras2007,Husson2019,Troyanskaya2001,Wang2022,Zhang2008}. For example, one representative approach involves constructing supervised machine learning models, such as linear regression, support vector machine, and random forests, that predict each variable as the objective variable using the remaining variables as explanatory variables~\cite{Bertsimas2018,Wang2022,Zhang2008}. Another approach is based on unsupervised learning and directly utilizes the similarity of observed data patterns among samples~\cite{Bras2007,Husson2019,Troyanskaya2001}. A typical example is $k$-nearest neighbors (k-NN), which estimates missing values by identifying similar samples based on their observed data~\cite{Bras2007,Troyanskaya2001}. Other examples include matrix factorization techniques, such as singular value decomposition, which approximate the data as a low-rank matrix, estimating missing values by uncovering latent structures and dependencies among variables~\cite{Husson2019}. Multiple imputation is another approach that integrates these individual imputation methods by generating multiple plausible datasets and combining the results through statistical pooling~\cite{rubin1996multiple}. This approach accounts for the uncertainty of missing data.

However, as discussed in Section~\ref{sec:intro}, these traditional imputation methods often struggle to achieve accurate imputation in real-world constructed-response tests, particularly in situations with high data sparsity and potential violations of the assumption of an underlying consistent data-generation process. This difficulty arises because they primarily infer missing values based on the relationships observed among the available scores (e.g., correlations between items or similarities between learners). When data sparsity is high, or when the assumption of a single underlying data-generation process is violated, these observed relationships become unreliable predictors for the missing scores, leading to biased or inaccurate imputations. To address these limitations, the main idea of this study is to leverage automated scoring technologies to impute missing scores.

\section{Automated Scoring for Constructed-Responses}

Recently, automated essay scoring and automated short-answer grading using artificial intelligence technologies have become prominent topics of artificial intelligence in the education community~\cite{ijcai2019-AesSurvey,Chaudhari2024,Kortemeyer2024,Li2024,Misgna2025,Uto2021bhmkAES}. While various methods have been proposed, conventional automated scoring approaches typically fall into one of two categories: feature-engineering-based approaches or neural-based approaches~\cite{ijcai2019-AesSurvey,Uto2021bhmkAES}.

Feature-engineering-based approaches rely on manually designed features, such as text length or the number of spelling errors, to predict scores using regression or classification models~\cite{Amorim2018AutomatedES,Burrows2015,AIED2017_AES,Leacock2003,AAAI1816447,AesBook2016}. While this approach offers interpretability and explainability, achieving high accuracy generally requires extensive effort in feature design and selection, which often needs to be tailored for each specific test item.

To address this limitation, neural-based approaches, which automatically extract features from data using deep neural networks, have gained increasing popularity. Early neural models have primarily employed convolutional neural networks or recurrent neural networks~\cite{Alikaniotis2016AutomaticTS,N18-1024,TDNN2018,mesgar-strube-2018-neural,D16-1193,wang-etal-2018-automatic}. More recent advancements have focused on using pretrained transformer networks~\cite{Transformer2017}, such as bidirectional encoder representations from transformers (BERT)~\cite{BERT2018}, which have demonstrated superior performance and accuracy in automated scoring tasks~\cite{AIED2019ShortAnswers,JiaqiBERT2020,mayfield-black-2020-fine,BertShortAnswer2019,Jin2021,yamaura2023}. BERT and its variants use extensive pretraining on large-scale text corpora, with high accuracy obtained by fine-tuning them for a target scoring task using relatively small datasets of scored responses.

Most recently, LLMs have emerged as the next frontier in automated scoring. LLMs build upon the transformer architecture, similar to BERT and its variants, but are pretrained on more massive and diverse datasets using various training techniques, such as reinforcement learning from human feedback and instruction tuning~\cite{Bai2022,Ouyang2022}. A major advantage of LLMs is their capability to address various natural language tasks, including automated scoring of constructed responses, by providing a concrete task explanation as a prompt in a zero-shot setting or by including a small number of examples alongside the prompt in a few-shot setting~\cite{Kojima2022}. This reduces the reliance on extensive labeled datasets, making LLMs highly adaptable to a wide range of natural language processing tasks. Recent studies exploring the application of LLMs for essay and short-answer scoring have shown that LLMs often achieve reasonable scoring performance using zero-shot or few-shot approaches~\cite{Kortemeyer2024,Misgna2025,chamieh2024llms,chang2024automatic,lee2024unleashing,mansour2024can,stahl2024exploring,wang2024beyond}, although they tend to perform worse than conventional fine-tuned scoring models based on pretrained transformers~\cite{chamieh2024llms,lee2024unleashing,mansour2024can,wang2024beyond,yancey2023rating}.

\section{Proposed Method}

This study proposes a novel imputation-based data augmentation approach using automated scoring technologies for accurate IRT-based ability estimation in constructed-response tests. The proposed approach consists of the following steps:
\begin{enumerate}
    \item \textbf{Developing a scoring model}: 
    Neural automated scoring models are prepared either by fine-tuning BERT or its variants on a subset of manually scored learner answers for each test item, or by employing a zero-shot scoring model using LLMs. The choice between these methods depends on various conditions as listed below:
    \begin{itemize}
        \item \textbf{Fine-tuned models}: These methods are recommended when a relatively large number of scored answers, such as more than a hundred, can be prepared for each item. This is often feasible in scenarios with a large number of examinees and when manual grading costs are acceptable.
        \item \textbf{Zero-shot scoring models}: These are more suitable when only a very limited number of scored answers are available for each item. For instance, this approach is preferable when the number of examinees is small, or when scoring individual answers is time-consuming. Zero-shot models are also suitable for situations in which clear scoring criteria are available or in which the evaluation task is relatively easy, because zero-shot evaluation is expected to be effective in such cases.
    \end{itemize}
    \item \textbf{Predicting missing scores}: 
    Once the scoring model is prepared, it is used to predict missing scores and construct a complete score dataset.
    \item \textbf{Estimating ability using IRT models}: 
    The augmented dataset is then used to estimate learner ability by applying IRT models.
\end{enumerate}

The proposed method is expected to achieve more robust imputation, even for heterogeneous data lacking a consistent data-generation process, because it directly leverages learner answer text to predict missing scores without modeling underlying patterns of score data. Additionally, it is expected to provide accurate imputations even under high missing ratios because constructing scoring models based on pre-trained neural language models or zero-shot models often achieves reasonable scoring performance with relatively few or no samples, as demonstrated in recent automated scoring studies~\cite{Kortemeyer2024,do2023,Jiang2024,Ridley2021AAAI,Shibata2024}. These features make the proposed method suitable for achieving accurate ability estimation while reducing the need for human grading.

\section{Experiments}

We conducted empirical evaluation experiments using real-world datasets to demonstrate the effectiveness of the proposed method.

\subsection{Data}

For our experiments, we required datasets comprising scored constructed responses for multiple items in which the same set of learners answered all items. Therefore, popular benchmark datasets for automated essay or short-answer grading tasks, such as ASAP~(automated student assessment prize)\footnote{https://www.kaggle.com/competitions/asap-aes} and ASAP--SAS~(short answer scoring)\footnote{https://github.com/benhamner/ASAP-SAS}, could not be used owing to the lack of information identifying respondents for each answer. Consequently, we utilized the following three datasets:
\begin{enumerate}
\item \textbf{Short-Answer Grading (SAG) Dataset}: This dataset, developed by the Benesse Educational Research and Development Institute in Japan, consists of responses from 511 Japanese university students to three short-answer items in a Japanese reading comprehension test. Scores for the responses were provided by expert raters using five rating categories for each item.
\item \textbf{Essay Scoring (ES) Dataset~\cite{Takeuchi2021}}: This dataset consists of essays written by 327 Japanese university students in response to three essay tasks offered in a natural science lecture. Each response was scored on a five-point scale by expert raters.
\item \textbf{ELYZA-tasks-100 (ELYZA) Dataset\footnote{https://huggingface.co/datasets/elyza/ELYZA-tasks-100/tree/main}}: This dataset is designed for evaluating Japanese LLMs. It includes responses generated by 33 LLMs for 100 writing tasks, with each response scored by expert raters using a five-point scale. We used this dataset by treating individual LLMs as learners and writing tasks as constructed-response test items. Although this dataset is not specifically intended for automated essay or short-answer scoring, it appears to be suitable for applying the proposed method with a zero-shot scoring model. This is because the number of examinees is small, and LLM-generated responses are relatively straightforward to score using LLMs in a zero-shot manner.
\end{enumerate}

\subsection{Experimental Procedures}

We conducted the following experiments for each dataset:
\begin{enumerate}
\item We estimated the IRT parameters from the complete score data based on the GPCM introduced earlier. The obtained ability estimates for learners were treated as gold-standard values in this experiment.
\item We created incomplete datasets from the complete score data by converting some scores into missing values. The missing ratios were varied as follows:  
\begin{itemize}
    \item For the SAG and ES datasets, we examined three missing ratios: 33\%, 50\%, and 62\%
    \item For the ELYZA dataset, we examined five missing ratios: 10\%, 20\%, 50\%, 65\%, and 80\%
\end{itemize}
The incomplete datasets were created following a systematic design~\cite{ihan2016,LinkingBook2014,Uto2020brm} to generate missing patterns while ensuring the applicability of IRT and the conventional missing imputation methods. The algorithm for creating missing patterns, which supports the rationale behind the selection of missing ratios for each dataset, is detailed in the Appendix.
\item Using each incomplete score dataset, we applied the GPCM to estimate learner ability through the following methods:
\begin{itemize}
    \item \textbf{Estimation without imputation}: This method directly applies the GPCM to the incomplete score dataset, ignoring missing scores during likelihood calculation as detailed in Section~\ref{sec:irt}.
    \item \textbf{Estimation with imputation by k-NN or random forest}: Ability is estimated by the GPCM after missing scores are imputed using k-NN or random forest~(RF)\footnote{We used the \texttt{VIM} and \texttt{missForest} libraries in R for missing score imputations using k-NN and RF, respectively.}, as introduced in Section~\ref{sec:imputation}.
    \item \textbf{Estimation with imputation by the proposed method}: Ability is estimated using the GPCM based on the proposed method. In the SAG and ES datasets, BERT-based models are used for the automated scoring models fine-tuned on scored answers for each item\footnote{We used BERT models with a linear output layer on top of the [CLS] token, which is appended to the beginning of the input text. The pre-trained BERT model was \texttt{tohoku-nlp/bert-base-japanese-v3}. The optimizer was AdamW with a learning rate of 1e-5. The mini-batch size was 64 for the SAG dataset and 16 for the ES dataset.}. For the ELYZA dataset, we employed a zero-shot scoring approach with GPT-4o, where the prompt was designed to include detailed task instructions with scoring criteria defined in the original dataset, as shown in Table~\ref{table:prompt}.
\end{itemize}
\item The root-mean-squared error (RMSE) and Pearson's correlation coefficient were calculated between the estimated and gold-standard abilities. To address the scale indeterminacy inherent in IRT estimation, the ability values were normalized to have a mean of zero and a variance of one, ensuring that the estimated and gold-standard abilities were directly comparable.
\item Steps 2 to 4 were repeated 10 times, each time using a different missing pattern. To create different missing patterns using the same algorithm, the order of learners was randomly shuffled with each repetition.
\end{enumerate}

\begin{table*}[t]
\caption{Translated prompt for zero-shot scoring on the ELYZA dataset. ($\dagger$)~indicates that remaining details have been abbreviated. The full Japanese grading criteria are available at \url{https://huggingface.co/datasets/elyza/ELYZA-tasks-100/blob/main/baseline/humaneval/guideline.md}.}
\label{table:prompt}
\centering
\begin{tabular}{p{148mm}}
\hline
You are a grader. You will be given a test item, a reference answer, a grading rubric, and a response. Referencing the grading rubric and the reference answer, grade the response on a scale of 1 to 5, and output the number only.\\
\textbf{Test Item:} \{text of the test item\}\\
\textbf{Reference Answer:} \{text of the reference answer\}\\
\textbf{Basic Grading Criteria}\\
{\it 1. Incorrect:} Does not follow instructions. Chooses an incorrect option in a multiple-choice question. ($\dagger$)\\
{\it 2. Incorrect but heading in the right direction:} Usually a 3-point answer with a 1-point deduction. ($\dagger$)\\
{\it 3. Partially correct:} Addresses the majority of a complex instruction correctly. ($\dagger$)\\
{\it 4. Correct:} Correctly answers the question. ($\dagger$)\\
{\it 5. Helpful:} Correctly answers the question and further anticipates user needs. ($\dagger$)\\
\textbf{Basic Deductions}: Scores may be adjusted based on the following factors.\\
{\it - Unnatural Japanese (-1 point):} Syntactically awkward or unclear Japanese, repetition of the same sentence, or abrupt insertion of English words.\\
{\it - Partial Hallucination (-1 point):} A response partially inconsistent with facts. ($\dagger$)\\
{\it - Excessive Safety Concerns (Score as 2 points):} e.g., Responds with ``I cannot answer for ethical reasons.''\\
\textbf{Item-Specific Grading Criteria:} \{Grading Criteria defined for each item\}\\
\textbf{Response:} \{Response to be evaluated\}\\
\hline
\end{tabular}
\end{table*}

Furthermore, we investigated the performance of the proposed method in situations in which the conventional approach was not applicable owing to a substantially high missing ratio. Specifically, we examined three additional missing ratios, namely, 70\%, 80\%, and 90\%, for the SAG and ES datasets, and a 100\% missing ratio for the ELYZA dataset in experimental procedure~2 for the proposed method. It should be noted that under these conditions, the conventional method cannot be applied due to the presence of learners with no assigned scores.

\subsection{Experimental Results}

\begin{figure}[t] 
\centering
\includegraphics[width=0.95\textwidth]{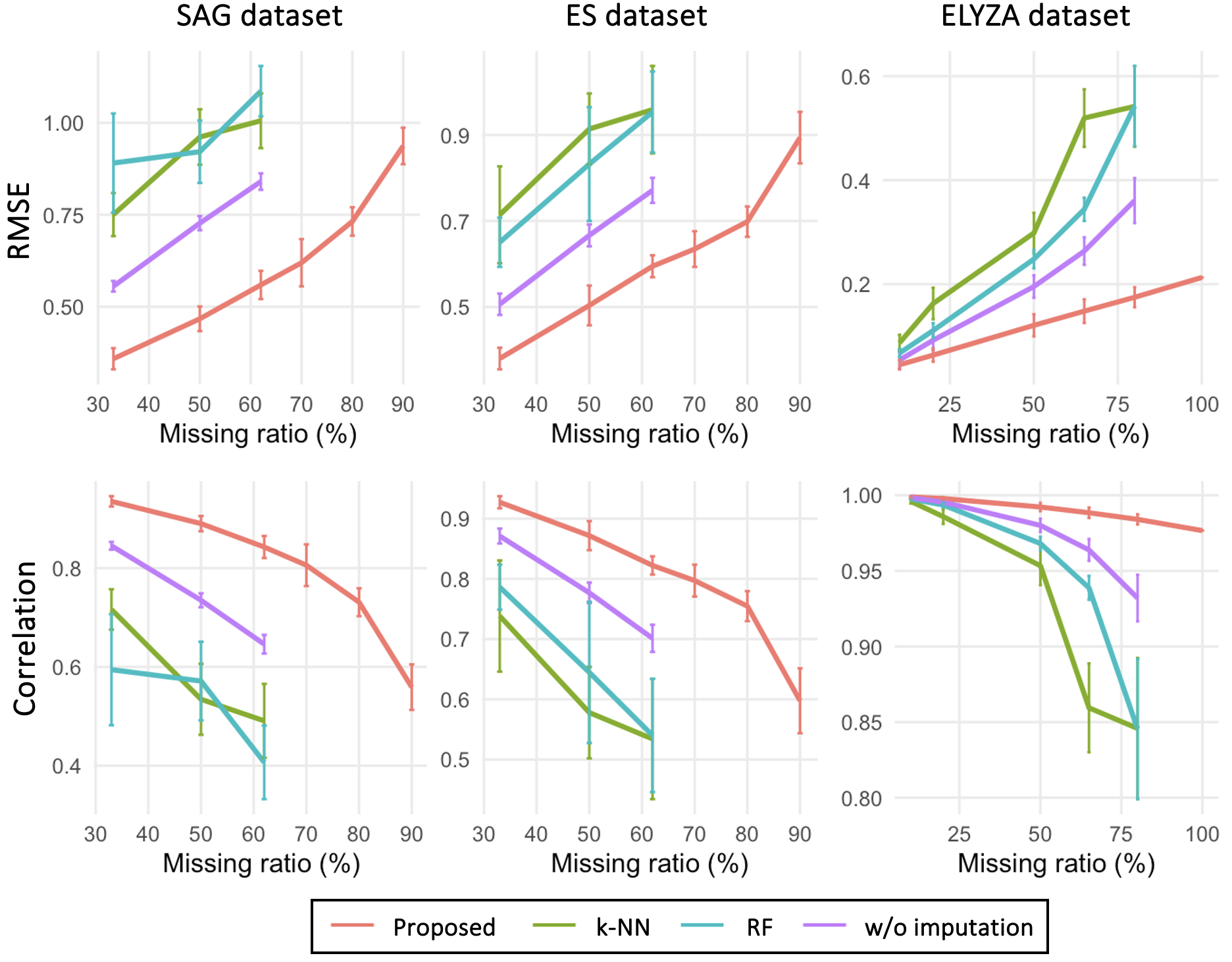}
\caption{Comparison of ability estimation accuracy across different missing ratios. Upper plots show RMSE and lower plots show Pearson's correlation coefficient between estimated and gold-standard abilities. Since the ability values are normalized to have zero mean and unit variance, an RMSE around 0.3 corresponds to about 5\% of the distribution range ($\pm$3 standard deviations), indicating high accuracy, while an RMSE around 0.6, corresponding to about 10\%, may still be acceptable.}\label{tab:main_results}
\end{figure}

Fig.~\ref{tab:main_results} shows the experimental results. The upper plot shows the averaged RMSEs, while the lower plot shows the averaged correlation coefficients, with error bars indicating the standard deviation over 10 repetitions. An exception is the condition with a 100\% missing ratio in the ELYZA dataset, in which results are based on a single trial due to the lack of variation in missing patterns. Note that the conventional methods have no results for missing ratios above 62\% in the SAG and ES datasets and 100\% in the ELYZA dataset because they are not applicable under these conditions, as described above.

According to the experimental results, the proposed method demonstrated lower RMSEs and higher correlation coefficients compared to conventional methods across all conditions. While all methods showed a reduction in the accuracy of ability estimation as the missing ratio increased, conventional imputation methods exhibited a more rapid decline in accuracy as the missing ratio increased than the proposed method, as highlighted in the results for the ELYZA dataset. This reflects the difficulty of conventional methods in modeling patterns underlying observed score data under high missing ratios. Additionally, conventional imputation methods underperformed compared to ability estimation without imputation, suggesting that inaccurate imputation introduces heavy bias into ability estimation.

A paired t-test was conducted to compare the average accuracy differences between the proposed method and each of the other methods at each missing ratio. The results indicated significant differences at the 1\% significance level for all comparisons, except for the comparison between the proposed method and the method without imputation at a missing ratio of 10\% in the ELYZA dataset. These findings indicate that the proposed method achieves remarkably high accuracy in ability estimation from incomplete data, except when the missing ratio is extremely low.

Furthermore, the proposed method achieved reasonable accuracy in ability estimation in situations in which conventional methods are not applicable. Specifically, it maintained high accuracy even with a 100\% missing ratio in the ELYZA dataset, corresponding to the case employing zero-shot scoring. The use of fine-tuned scoring models in the SAG and ES datasets also demonstrated relatively good accuracy with a missing ratio of up to 80\%. More specifically, the accuracies under the 80\% missing ratio were higher than those of conventional methods at a 62\% missing ratio. These results highlight the unique and important advantage of the proposed method for accurately estimating abilities, except in cases of extremely large missing ratios, such as at a missing ratio of 90\%, when fine-tuned scoring models are used.

\section{Analysis}

\subsection{Accuracy of Missing Score Prediction}

\begin{figure}[t] 
\centering
\includegraphics[width=0.95\textwidth]{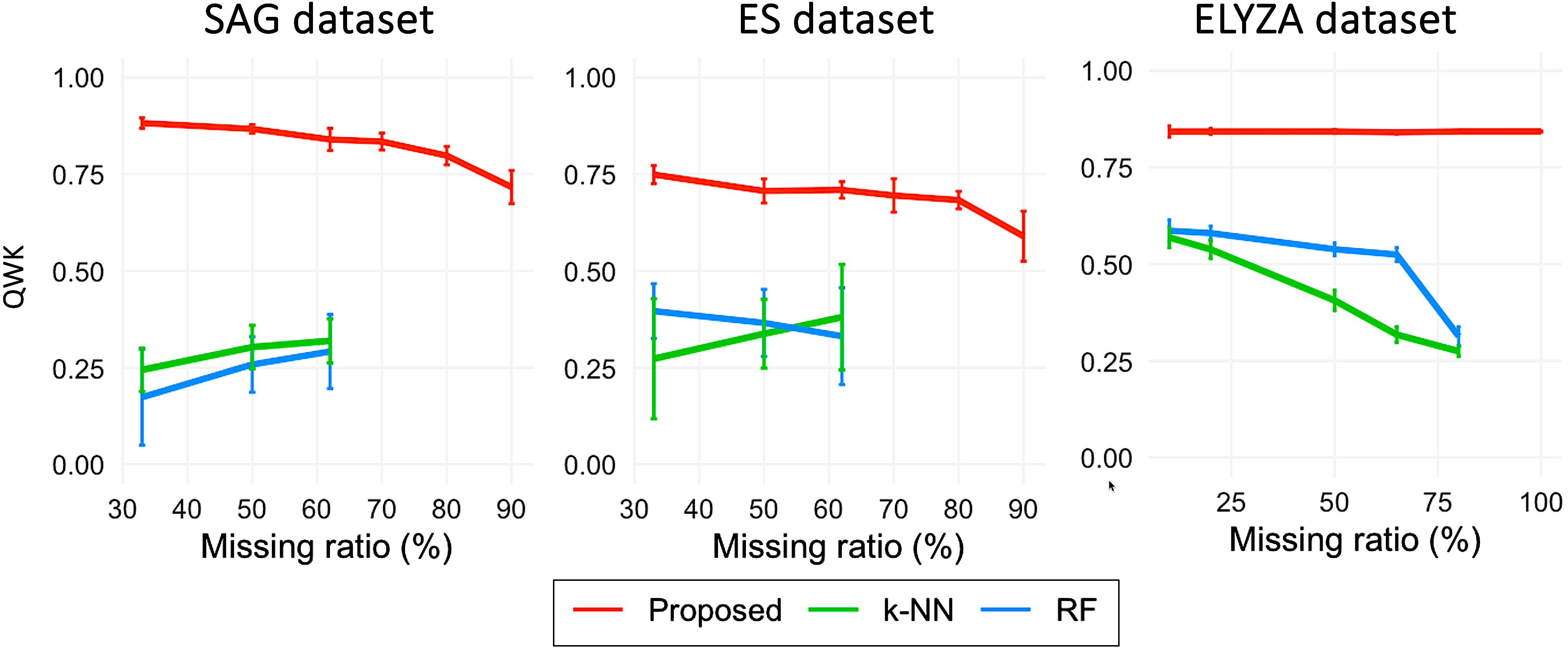}
\caption{Accuracy of predicting missing scores for each imputation method, measured by QWK.}\label{fig:acc}
\end{figure}

We can infer that the high accuracy in ability estimation of the proposed method comes from its high accuracy in missing score imputation. To confirm this, we analyzed the accuracy of missing score imputation for each method. Specifically, using the results from experimental procedure~3, we calculated the agreement between the predicted scores for missing values generated by each imputation method, including k-NN, RF, and the proposed method, and their corresponding true scores. As an evaluation metric, we used quadratic weighted kappa (QWK), which is commonly employed in research on automated scoring. 

Fig.~\ref{fig:acc} shows the results, with the average QWK values with error bars representing the standard deviations obtained from 10 repeated experiments. Given that higher QWK values indicate greater imputation accuracy, the results show that the proposed method achieves high accuracy in imputing missing scores, whereas conventional methods exhibit markedly lower accuracy.

Furthermore, in the SAG and ES datasets, the proposed method exhibits a decline in accuracy as the missing ratio increases due to the reduction in training data for fine-tuning, while the degraded accuracies remain higher than those of conventional methods. Moreover, in the ELYZA dataset, while conventional methods suffer a drastic decline in imputation accuracy, the proposed method maintains high accuracy regardless of the missing ratio. These results suggest that the proposed method achieves high imputation accuracy, which is likely to contribute to its high ability estimation accuracy.

\subsection{Robustness to Heterogeneity}

As described in Section~\ref{sec:intro}, the proposed method is expected to be effective for imputing missing scores that are difficult to predict from the patterns of observed data owing to their heterogeneity. To demonstrate this, Table~\ref{table:results3} provides examples of score imputations by the proposed method and the k-NN method for two learners sampled from the results of the SAG dataset. 

In the table, the {\it True scores} row represents the original complete data and the {\it w/o imput.} row shows the incomplete data (with missing values indicated by ``NA''). The {\it k-NN} and {\it Proposed} rows show the complete data created by imputing the missing values using each respective method. Additionally, the {\it True avg.} row indicates the averaged scores for each item calculated from the complete data, while $\hat{\theta}$ represents the estimated abilities based on each score dataset.

\begin{table}[t]
\centering
\caption{Examples of score imputation by the k-NN and proposed methods.}\label{table:results3}
\begin{minipage}{0.48\linewidth}
\centering
\begin{tabular}{c|ccc|c}\Xhline{1pt}
 & Item 1 & Item 2 & Item 3 & $\hat{\theta}$ \\
 \hline
True scores & 4 & 5 & 3 & 0.35 \\
w/o imput. & 4 & 5 & NA & 1.77 \\
k-NN & 4 & 5 & 5 & 2.35 \\
Proposed & 4 & 5 & 2 & 0.09 \\
\hline
True avg.  & 2.31 & 3.22 & 3.84 &  \\
\Xhline{1pt}
\end{tabular}
\end{minipage}
\hfill
\begin{minipage}{0.48\linewidth}
\centering
\begin{tabular}{c|ccc|c}\Xhline{1pt}
 & Item 1 & Item 2 & Item 3 & $\hat{\theta}$ \\
 \hline
True scores & 2 & 5 & 3 & -0.25 \\
w/o imput. & NA & NA & 3 & -0.84 \\
k-NN & 2 & 3 & 3 & -0.63 \\
Proposed & 2 & 5 & 3 & -0.12 \\
\hline
True avg.  & 2.31 & 3.22 & 3.84 &  \\
\Xhline{1pt}
\end{tabular}
\end{minipage}
\end{table}

Considering that the average item scores follow the order of item 3 $>$ item 2 $>$ item 1, predicting the missing data in the left table as a score of 5, as done by the k-NN method, might seem reasonable. Similarly, in the right table, given that the score for item 3 was 3, predictions by the k-NN method for the two missing data items (item 1, 2; item 2, 3) appear reasonable based on the overall trend. However, these predictions by the k-NN method are quite inaccurate. This discrepancy arises because the true scores in these cases do not follow the surrounding patterns, making them difficult to predict.

As shown in the examples above, conventional methods struggle to handle such data. In contrast, the proposed method does not rely on modeling surrounding score patterns but instead evaluates the content of individual answer texts. Therefore, the proposed method can make accurate predictions for heterogeneous score data.

\section{Conclusion}

This study proposed a novel method for imputing missing scores to enhance IRT-based ability estimation by leveraging automated scoring technologies. Experimental results demonstrated that the proposed method achieves higher accuracy in ability estimation compared with conventional approaches, even under conditions of high missing ratios or heterogeneous data. This indicates that the proposed method achieves high accuracy in ability estimation while markedly reducing the manual grading workload. 

However, we acknowledge several limitations. Firstly, the effectiveness of our method depends on the accuracy of the employed automated scoring model. Inaccurate scoring models, whether fine-tuned on insufficient data or based on zero-shot LLMs with suboptimal prompts, could lead to biased imputations and consequently affect the validity of the final ability estimates. The observed accuracy drop at the 90\% missing ratio when using fine-tuned models might partially reflect this dependency. Secondly, developing these scoring models can require substantial computational resources for fine-tuning or careful prompt engineering for zero-shot approaches, potentially offsetting the intended reduction in manual grading workload. Thirdly, while we demonstrated effectiveness on specific datasets, all datasets were limited to the Japanese language, which restricts the generalizability of our findings to broader linguistic and assessment contexts. Fourthly, the proposed method treats imputed scores as deterministic inputs to IRT estimation and does not propagate the uncertainty associated with automated scoring, which could result in overconfident or biased ability estimates. Lastly, the use of black-box models such as BERT and LLMs limits the interpretability of individual imputed scores and hinders systematic fairness evaluation, an important consideration for high-stakes testing scenarios.

Future research should address these limitations and explore further extensions. Although this study focused on unidimensional IRT, the proposed method may be applicable in various ability measurement contexts. For instance, the method is applicable to complex student models, such as multidimensional IRT~\cite{Reckase2009}, cognitive diagnostic models~\cite{templin2006measurement}, and knowledge tracing~\cite{Shen2024}. These complex models are more sensitive to data sparsity, which may further highlight the advantages of the proposed approach. Furthermore, exploring methods to incorporate the uncertainty associated with automated scores could lead to more reliable ability estimates and associated error measures. In addition, methods to enhance the interpretability of imputed scores should be explored to increase transparency and trust in the system. Evaluating the method on more diverse datasets across multiple languages, domains, and assessment formats is also a key direction for improving external validity. In addition, future studies should consider comparing the proposed method with more advanced deep-learning-based imputation methods, such as deep matrix factorization~\cite{xue2017deep} and variational autoencoders, which were not included in the current evaluation. Another promising direction is to leverage IRT models with rater parameters~\cite{Uto2020bhmk,MFRM2016BOOK,Uto2021bhmkIRT,Uto2024Pros}, such as many-facet Rasch models, to achieve more valid ability estimation by treating AI graders as distinct raters within a unified measurement framework. Finally, studying the integration of this method into practical applications like adaptive testing systems for constructed-response items, where real-time scoring and imputation could enhance test efficiency and personalization, would be valuable. Given the critical role of accurate ability estimation in learning support systems, we believe these future investigations hold significant promise.

\section*{Appendix: Algorithms for Generating Missing Patterns}

In this appendix, we explain how incomplete data were created from complete data in the experiments. For the missing ratios of 33\%, 50\%, and 62\% in the SAG and ES datasets, the missing patterns, which enable IRT parameter linking~\cite{ihan2016,LinkingBook2014,Uto2020brm} and the application of other imputation methods, were generated as follows:
\begin{itemize}
    \item The 33\% missing pattern was created by repeating the patterns of learners 1--3 on the left side of Table~\ref{table:ex_design}.
    \item The 50\% missing pattern was created by repeating the patterns of learners 1--6 on the right side of Table~\ref{table:ex_design}.
    \item The 62\% missing pattern was created by combining one repetition of learners 1--3 with six repetitions of learners 4--6, based on the pattern on the right side of Table~\ref{table:ex_design}.
\end{itemize}

For the missing ratios of 10\%, 20\%, 50\%, 65\%, and 80\% in the ELYZA dataset, these missing patterns were generated using Algorithm~1, which also ensures parameter linking in IRT estimation and the application of other imputation methods

\begin{algorithm}[h]
\caption{Algorithm for creating missing patterns for the ELYZA dataset.}\label{alg3} 
\begin{algorithmic} 
\State{Input: ${\cal I}$, ${\cal J}$, $N_m$: Number of missing ratio}
\State{Initialize missing indicator variable $\{ z_{ij} = 1 \mid i \in {\cal I}, j \in {\cal J}\}$ defined in Equation (3)}
\For{$j \in {\cal J}$}
  \State{Set $i_{st} = (j*10) \% 100$ and $i_{ed} = (j*10 + N_m) \% 100$}
  \If{$i_{ed} > i_{st}$}
    \State{Set $z_{ij} = 0$ for $i$ in $[i_{st}:i_{ed}]$} 
  \Else
    \State{Set $z_{ij} = 0$ for $i$ in $[i_{ed}:100]$ and $[1:i_{st}]$}    
  \EndIf	
\EndFor
\end{algorithmic}
\end{algorithm}

For the other missing ratios in the SAG and ES datasets, where IRT and conventional imputation methods are not applicable, missing patterns were generated by randomly selecting learners for each item until the specified percentage of scores was converted to missing values.

\begin{acknowledgments}
This work was supported by JSPS KAKENHI Grant Numbers 23K20727 and 24H00739. We thank Yuki Doka and Yoshihiro Kato from the Benesse Educational Research and Development Institute for permission to use the SAG dataset. 
\end{acknowledgments}

\section*{Declaration on Generative AI}
During the preparation of this work, the authors used GPT-4o in order to: Grammar and spelling check. After using this tool, the authors reviewed and edited the content as needed and take full responsibility for the publication’s content. 

\bibliography{extracted.bib}

\end{document}